\documentclass[runningheads]{llncs}
\usepackage{appendix}
\usepackage{hyperref}
\usepackage{amssymb}
\usepackage[T1]{fontenc}
\usepackage{multirow}
\usepackage{booktabs}
\usepackage{makecell}
\usepackage{graphicx,verbatim}
\begin{document}
\title{CRUNet-MR-Univ: A Foundation Model for Diverse Cardiac MRI Reconstruction}
%
\begin{comment}  %% Removed for anonymized MICCAI 2025 submission
\author{First Author\inst{1}\orcidID{0000-1111-2222-3333} \and
Second Author\inst{2,3}\orcidID{1111-2222-3333-4444} \and
Third Author\inst{3}\orcidID{2222--3333-4444-5555}}
%
\authorrunning{F. Author et al.}
\institute{Princeton University, Princeton NJ 08544, USA \and
Springer Heidelberg, Tiergartenstr. 17, 69121 Heidelberg, Germany
\email{lncs@springer.com}\\
\url{http://www.springer.com/gp/computer-science/lncs} \and
ABC Institute, Rupert-Karls-University Heidelberg, Heidelberg, Germany\\
\email{\{abc,lncs\}@uni-heidelberg.de}}

\end{comment}

% \author{Anonymized Authors}  %% Added for anonymized MICCAI 2025 submission
% \authorrunning{Anonymized Author et al.}
% \institute{Anonymized Affiliations \\
%     \email{email@anonymized.com}}

               % typeset the header of the contribution
%
\author{Donghang Lyu\inst{1} \and Marius Staring\inst{1} \and Hildo J. Lamb\inst{1} \and Mariya Doneva\inst{2}}
% \author{Anonymized Authors}  %% Added for anonymized MICCAI 2025 submission

\institute{Department of Radiology, Leiden University Medical Center, Leiden, The Netherlands\\ 
\email{d.lyu@lumc.nl}
% \email{\{d.lyu, m.staring, m.j.p.van\_osch, h.j.lamb\}@lumc.nl}\\
\and
Philips Innovative Technologies, Hamburg, Germany}

\maketitle   
% making it challenging for a single deep learning method to generalize across all scenarios. 

\begin{abstract}
In recent years, deep learning has attracted increasing attention in the field of Cardiac MRI (CMR) reconstruction due to its superior performance over traditional methods, particularly in handling higher acceleration factors, highlighting its potential for real-world clinical applications. However, current deep learning methods remain limited in generalizability. CMR scans exhibit wide variability in image contrast, sampling patterns, scanner vendors, anatomical structures, and disease types. Most existing models are designed to handle only a single or narrow subset of these variations, leading to performance degradation when faced with distribution shifts. Therefore, it is beneficial to develop a unified model capable of generalizing across diverse CMR scenarios. To this end, we propose CRUNet-MR-Univ, a foundation model that leverages spatio-temporal correlations and prompt-based priors to effectively handle the full diversity of CMR scans. Our approach consistently outperforms baseline methods across a wide range of settings, highlighting its effectiveness and promise.

\keywords{Diverse CMR Reconstruction \and Foundation Model \and Prompt}
% Authors must provide keywords and are not allowed to remove this Keyword section.

\end{abstract}
\section{Introduction}
Cardiac MRI (CMR) is widely used in clinical practice for assessing cardiovascular function, offering high-resolution images and excellent soft tissue contrast. To shorten long acquisition times and reduce breath-hold discomfort, undersampling is commonly used to accelerate scanning. CMR reconstruction then restores the image from the undersampled k-space data, which involves reducing artifacts and noise. Compared to traditional methods like Parallel Imaging~\cite{1,2} and Compressed Sensing~\cite{3}, deep learning methods~\cite{4,5,6,7,8} for CMR reconstruction are gaining attention for their stronger performance at higher acceleration factors.

Despite these advancements, a lot of deep learning approaches remain constrained to specific scenarios, largely due to the use of highly specialized training data. This results in performance degradation when applied to data with different distributions, which is common in practice, where variations span image contrast, sampling trajectories, scanner vendors, anatomical structures, and disease types, etc. Given these diversities, training a separate model for each specific scenario is impractical, underscoring the need for a reconstruction foundation model that generalizes across diverse CMR settings. Since the introduction of the GPT series~\cite{9,10,11}, foundation models have gained traction due to their ability to learn from large, diverse datasets and generalize across tasks. In the medical domain, numerous foundation models have emerged, such as MedSAM models~\cite{12,22} for medical image segmentation and vision-language models~\cite{13,14} for report generation and visual question answering. Although recent efforts from the CMRxRecon 2024 challenge~\cite{8} have aimed to address variability across contrast, acceleration factor, and sampling pattern, they still fall short of capturing the full diversity of real-world CMR scans.

In this work, we propose \textbf{CRUNet-MR-Univ}, a foundation model designed for diverse CMR reconstruction that combines an unrolled architecture with Convolutional Recurrent U-Net (CRUNet) model~\cite{25} and prompt-based priors to enhance generalization. Recognizing the inherent temporal dimension in most CMR scans, our model leverages rich spatio-temporal information across the entire sequence. Unlike CRNN-MRI~\cite{4}, which employs a basic convolutional recurrent design, CRUNet integrates bidirectional recurrence into a U-Net by splitting it into two unidirectional units with opposite directions, placed separately in the encoder and decoder. This enables continuous spatio-temporal feature extraction. Additionally, inspired by previous methods such as PromptMR~\cite{5}, PCP-UNet~\cite{15}, and UPCMR~\cite{16}, we incorporate both learnable and text-based prompts to encode diverse CMR scan attributes and help improve robustness. After training and evaluating on the CMRxRecon2025 dataset\footnote{\url{https://www.synapse.org/Synapse:syn59814210/wiki/631023}} , which includes data from multiple medical centers, scanner vendors, field strengths, disease types, image contrasts, sampling trajectories, and acceleration factors, CRUNet-MR-Univ demonstrates stronger performance over the other baseline methods.

\section{Methodology}
\subsection{CRUNet-MR-Univ}
\begin{figure}[tb]
\centering
\includegraphics[width=\textwidth]{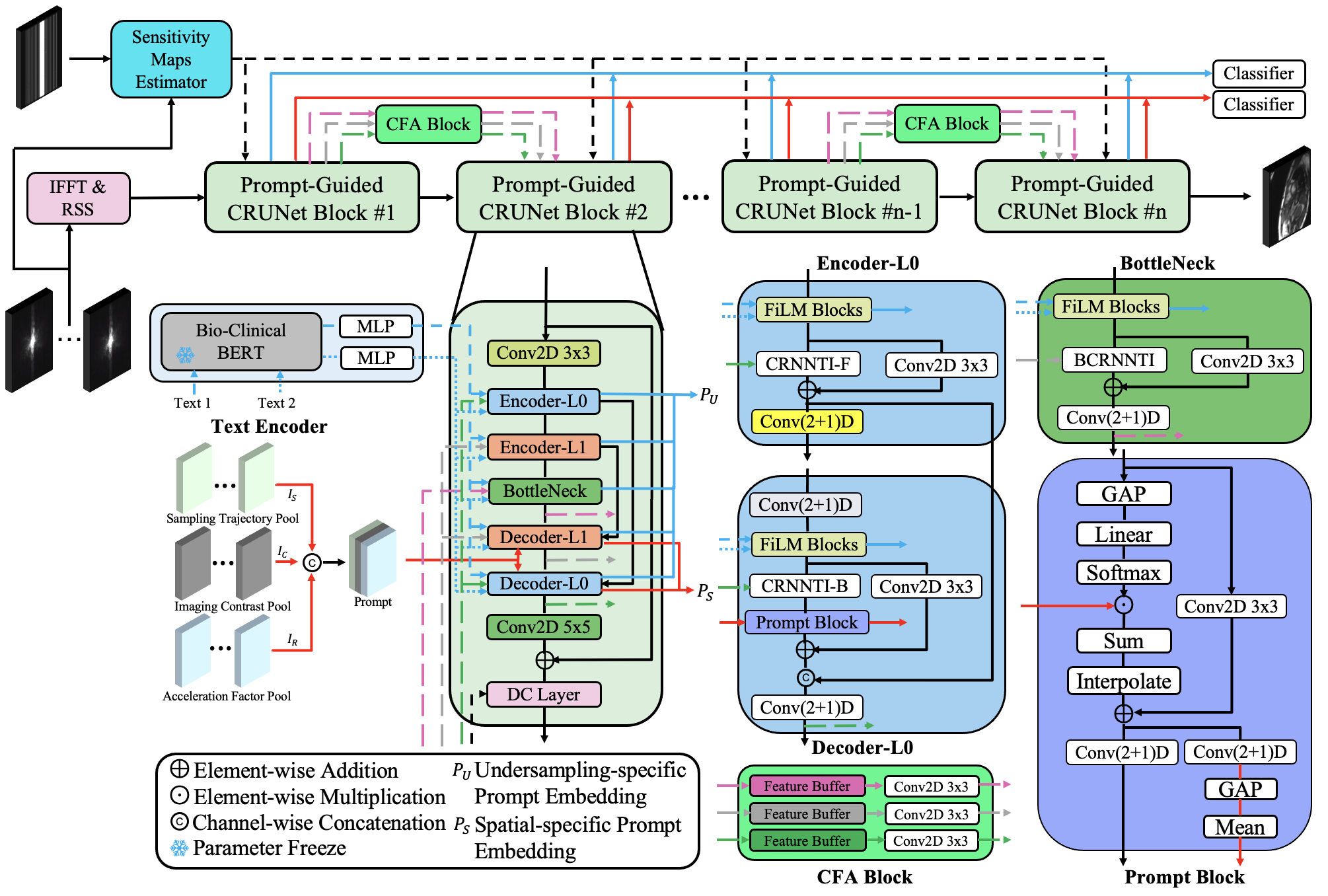}
\caption{Overview of the CRUNet-MR-Univ model. The bottom section details the structure of each cascade block and its core components. Each cascade block contains a CRUNet model, with pink, gray, and green long-dashed lines showing the flow of hidden state features within the CRNNTI block at each level. Blue dotted and dashed lines represent text prompt inputs, while blue and red solid lines denote the flow of output undersampling-specific and spatial-specific prompt embeddings, respectively. The black dashed lines correspond to the input of the estimated CSM.} \label{fig1}
\end{figure}

Overall, CRUNet-MR-Univ adopts an unrolled network design due to the iterative nature of MRI reconstruction, as illustrated in Figure~\ref{fig1}. From a global perspective, the model takes two inputs: an undersampled multi-coil k-space and its corresponding sampling mask. Firstly, the k-space data is transformed into the image domain using an inverse Fourier transform. A coil-combined image is then generated via the Root Sum of Squares (RSS) operation across coils, which serves as the input to the following cascade block. Simultaneously, a temporally averaged autocalibration signal (ACS) region, derived based on the mask, is fed into a Sensitivity Maps Estimator (SME) block, adapted from PromptMR~\cite{5}, which estimates coil sensitivity maps (CSMs). Furthermore, to promote effective information flow across cascades, a Cascaded Feature Aggregation (CFA) block is introduced to aggregate all preceding feature maps to guide the convolutional recurrent modules in subsequent cascade. Following the prompt design in UPCMR~\cite{16}, each cascade block also incorporates two kinds of prompts: an undersampling-specific prompt $P_U$ and a spatial-specific prompt $P_S$. These prompts interact with image features to generate joint embeddings, which are concatenated across cascades and used for classifications to condition the reconstruction process. Specifically, separate MLP-based classifiers are assigned to each prompt type, predicting imaging contrast, sampling trajectory, and acceleration factor. This design encourages each prompt to better capture its associated contextual information.
\\
\\
\textbf{Prompt-guided CRUNet Block} 
CRUNet~\cite{25} builds upon the CRNN-MRI~\cite{4} model, a simple yet effective network for cine MRI reconstruction that leverages the strong spatio-temporal correlations within the sequence. CRNN-MRI incorporates two types of convolutional recurrent units: Bidirectional Convolutional Recurrent Units evolving over Time and Iterations (BCRNNTI), and Convolutional Recurrent Units evolving over Iterations (CRNNI). The former enables information propagation both across the temporal sequence and between cascade blocks, while the latter focuses solely on iterative refinement across cascade blocks. However, in its original design, only a single BCRNNTI block is placed at the beginning of each cascade block, which limits the continuous extraction of spatio-temporal features throughout the cine sequence. This discontinuity can hinder the model's ability to capture internal spatio-temporal features effectively. Therefore, CRUNet enhances the architecture by splitting a BCRNNTI block into two CRNNTI blocks with forward and backward propagation directions (i.e., CRNNTI-F and CRNNTI-B) and placing them in the encoder and decoder parts of U-Net structure, respectively. Their outputs are fused via skip connections, effectively forming an enhanced BCRNNTI unit that captures temporal information from neighboring frames while integrating both low-level and high-level spatial features. As shown in Figure~\ref{fig1}, each CRUNet follows a two-level U-Net structure with two convolutional layers at each end for adjusting the channel number, which is fixed at 64 throughout the model. The two intermediate levels share a similar design, except that Conv(2+1)D layers are used for downsampling and upsampling in the first level. A full BCRNNTI block is placed at the bottleneck to keep extracting spatio-temporal features within the sequence. Furthermore, to enhance spatial feature extraction via a larger receptive field, we apply dilation factors of (1, 2, 4) to the two levels and bottleneck of CRUNet. These are used in both the CRNNTI units and Conv(2+1)D layers.

At each level of the encoder and decoder, we incorporate two types of prompts, each providing a distinct perspective. To leverage richer prior information, we design text-based prompts consisting of two components: one encoding scanner-specific information (vendor, model, field strength), and the other capturing CMR acquisition details (contrast, sampling trajectory, acceleration factor). These are formatted using two templates: (1) \textit{"\{vendor\} \{model\} MRI scanner at \{field strength\} field strength"} and (2) \textit{"MRI scan of \{contrast\}, sampled using \{sampling trajectory\} trajectory with an acceleration factor of \{acceleration factor\}"}. The textual prompts are processed by a frozen Bio-Clinical BERT~\cite{17}, pretrained on a large-scale clinical corpora, and subsequently refined via MLPs to match the target dimensionality. For each encoder and decoder block, the resulting prompt embeddings are integrated using two FiLM blocks~\cite{18}. Each FiLM block generates modulation parameters, weight $W_{P} \in \mathbb{R}^{B \times T \times C}$ and bias embeddings $B_{P} \in \mathbb{R}^{B \times T \times C}$, by taking the concatenation of prompt and feature embeddings as input. Here, $B$ denotes the batch size, $T$ the number of frames, and $C$ the number of channels. These parameters modulate the feature representations, enabling dynamic conditioning on prior information. It is worth noting that, in the second FiLM block, $W_{P}$ and $B_{P}$ are summed and averaged across the temporal dimension to obtain an updated undersampling-specific prompt embedding.

In the decoder part, we introduce a PromptBlock with a learnable spatial-specific prompt $P_{S} \in \mathbb{R}^{3 \times C \times H \times W}$, where $H$ and $W$ denote the initial spatial size. $P_S$ is formed by concatenating three learnable embeddings, each selected from a prompt pool based on prior information: sampling trajectory ($I_{S}$), imaging contrast ($I_{C}$), and acceleration factor ($I_{R}$). Following the prompt design in PromptIR~\cite{19}, a weight embedding is generated from the input feature map to modulate the first dimension of $P_{S}$ via a weighted sum, effectively fusing prior information. The resulting prompt is then interpolated and added to the input feature map, enhancing it with the integrated priors. An updated spatial-specific prompt embedding is further obtained through global average pooling followed by temporal averaging.
\\
\\
\textbf{CFA Block} In the CRNNTI block, information is propagated across cascade blocks by treating the output feature map from the previous cascade as an additional input for the current one. However, as the network becomes deeper, information from earlier cascade blocks tends to highly degrade or vanish, despite potentially containing useful context for the current cascade. To address this, a Cascaded Feature Aggregation (CFA) block is introduced. Given the $j$-th level of the $i$-th cascade block, we maintain a feature buffer that stores all previous feature maps at that level, denoted as $F_{j,0}, \dots, F_{j,i-1}$. These feature maps are concatenated along the channel dimension and passed through a convolutional layer to produce an updated feature map, effectively integrating information from all preceding cascade blocks. Notably, CRUNet comprises two levels and a bottleneck, yielding three feature buffers in the CFA block, one for each of them, with dilation factors of convolution layers matching those used in CRUNet.

\subsection{Loss Function}
The overall loss function is composed of two parts: reconstruction loss $\mathcal{L}_{rec}$ and classification loss $\mathcal{L}_{cls}$. The $\mathcal{L}_{cls}$ is the sum of the cross-entropy losses for the contrast class, sampling trajectory class and acceleration factor class, while the $\mathcal{L}_{rec}$ is the weighted sum of L1, MSE and SSIM loss terms, defined as follows:
\begin{equation}
    \mathcal{L}_{rec} = \lambda_{l1} \| |I_{rec}| - |I_{gnd}| \|_1 + \lambda_{l2} \| |I_{rec}| - |I_{gnd}| \|_2^2 + \lambda_{ssim}(1-\mathrm{SSIM}(|I_{rec}|, |I_{gnd}|)),
\end{equation}
where $I_{rec}$ denotes the reconstructed CMR image sequence and $I_{gnd}$ represents the ground-truth sequence. All loss terms are computed using the absolute value of the images. We set $\lambda_{l1}=\lambda_{l2}=0.5$ and  $\lambda_{ssim}=1$. Finally, the overall loss function represents as follows:
\begin{equation}
    \mathcal{L} = \lambda_{cls}\mathcal{L}_{cls}+\mathcal{L}_{rec}.
\end{equation}
Since classification serves as an auxiliary task primarily for guiding prompt tuning, which is much less important than the main reconstruction objective, we assign it a small weight of $\lambda_{cls} = 0.025$.

\section{Experimental Setup}
\subsection{Dataset and Task Description}
The CMRxRecon2025 challenge aggregates data from over 5 medical centers and more than 10 MRI scanners from GE, Philips, Siemens, and United Imaging,  including 1.5T and 3.0T scans. Furthermore, the dataset spans multiple MRI modalities and sequences: bSSFP is used for cine, phase-contrast (PC), and tagging sequences; FLASH is employed for mapping and dark-blood imaging; and TSE is utilized for T2-weighted imaging. Likewise, the dataset encompasses a variety of cardiac diseases. The dataset comprises multi-parametric CMR imaging from 600 subjects, divided into 200 training, 100 validation, and 300 testing cases. Consequently, the challenge focuses on developing a foundation model that generalizes well to unseen data from different medical centers and across diverse cardiovascular diseases. In this paper, we focus on the first task, evaluating model generalization across multiple centers.

Within the training dataset, three acceleration factors (8×, 16×, 24×) and three sampling patterns (uniform Cartesian, Gaussian Cartesian, pseudo-radial) are provided. Notably, Gaussian Cartesian and pseudo-radial trajectories employ temporal interleaving, whereas uniform Cartesian does not. The ACS region comprises the central 20 lines for Cartesian trajectories and a 20×20 central area for pseudo-radial sampling.

\subsection{Implementation Details}
The training of CRUNet-MR-Univ was conducted in two stages. In the first stage, we verified the effectiveness of key components (i.e., the CFA block and prompt modules). In the second stage, we further unlocked the model’s potential by adjusting training settings and employing a curriculum learning strategy.

\subsubsection{The First Stage}
Before training, several preprocessing steps were applied. To accommodate CRUNet’s requirement for temporal input, modalities without a time dimension (e.g., black-blood, T1w, T2w) were expanded into single-frame sequences. Although CRUNet supports variable frame counts, some samples with excessive frames (e.g., 54) hindered recurrent operations and highly increased computational cost. Therefore, for cases with more than 12 frames, we randomly selected 12 continuous frames for training. Furthermore, to address the imbalance in the number of training samples across modalities, we enforced uniform sampling in the data loader, ensuring roughly equal exposure per modality in each training epoch. For data normalization, we transformed the multi-coil k-space data into the image domain, normalized it by dividing by the maximum absolute value, and then converted it back to the k-space domain.

The models were implemented in PyTorch 2.0.0 and trained on an NVIDIA A100 GPU with 80GB memory. To optimize GPU usage, we employed mixed precision training~\cite{20}. To enable faster evaluation and reduce overall training time, all unrolled methods used 6 cascade blocks and were trained for 60 epochs, with 6,000 samples selected per epoch. Batch size was set to 1. We used the AdamW optimizer with parameters $\beta_{1}=0.9$, $\beta_{2}=0.999$, $\epsilon=10^{-8}$, an initial learning rate of $2 \times 10^{-4}$ and a weight decay of 0.1. The learning rate was reduced by a factor of 0.9 every two epochs, with a minimum threshold of $2 \times 10^{-5}$. In addition to CRUNet-MR-Univ, we evaluated two additional related baseline methods for comparison. The first is CRNN-MRI~\cite{4}, and the second adopts the CRUNet architecture but replaces all CRNNTI blocks with standard Conv3D blocks (i.e. UNet-MR). Notably, both models did not incorporate prompts.

\subsubsection{The Second Stage} We modified the preprocessing procedure to improve inference on longer sequences and reduce training overhead, as randomly selecting 12 consecutive frames in our earlier setup hindered reconstruction performance of longer sequences. Inspired by previous studies~\cite{5,24}, we adopted a strategy of using five consecutive frames as input while focusing the reconstruction on the middle frame. For cases with fewer than five frames, model was configured to output the entire sequence directly. This approach enhances inference by reconstructing each frame through the exploitation of spatio-temporal correlations with neighboring frames, while simultaneously reducing GPU memory consumption, thereby enabling the use of more cascade blocks in CRUNet-MR-Univ during training.

Another implementation change is the training strategy: we adopted curriculum learning~\cite{21} to enable progressive, step-wise learning in CRUNet-MR-Univ. The details are as follows:
\begin{enumerate}
    \item Initialize the model with 6 cascade blocks and train for 40 epochs with an acceleration factor of \{8\}.
    \item Add 4 new blocks to the model (total 10 blocks), train for 40 epochs with an acceleration factor of \{8, 16\}, with the sampling probabilities of \{0.2, 0.8\}.
    \item Add 2 new blocks to the model (total 12 blocks), train for 32 epochs with an acceleration factor of \{8, 16, 24\}, with the sampling probabilities of \{0.1, 0.1, 0.8\}.
    \item Train the complete model for 13 epochs using all the acceleration factors with the equal sampling probability.
\end{enumerate}
For the first three steps, each epoch was trained with 6000 samples, while the final step used 16000 samples. The first three steps employed a cosine-annealing scheduler with warm-up. Steps one and two used 6 warm-up epochs with learning rates of $2 \times 10^{-4}$ and $1 \times 10^{-4}$, and a minimum learning rate of $1 \times 10^{-5}$. Step three used 5 warm-up epochs, a learning rate of $5 \times 10^{-5}$, and a minimum learning rate of $1 \times 10^{-6}$. In the final step, the initial learning rate was set to $8 \times 10^{-5}$ and reduced by a factor of 0.4 every two epochs, then it was set to $1 \times 10^{-7}$ in the last epoch.
\\
\\
For evaluation metrics, peak signal-to-noise ratio (PSNR), structural similarity index (SSIM), and normalized mean squared error (NMSE) were chosen, computed on the cropped central region of each validation case via submission to the challenge website.

\section{Results}
The organizers have evaluated several traditional methods such as SENSE, GRAPPA, and zero-filled (ZF) reconstruction. Then Table~\ref{tab1} summarizes the overall performance comparison, covering both baseline methods and ablation studies. Our proposed CRUNet-MR-Univ consistently outperforms related baselines on the small center-cropped region. The ablation results further confirm the positive contributions of the CFA blocks and prompt components to overall performance. Comparing the two training stages of CRUNet-MR-Univ reveals that introducing additional cascade blocks in combination with a curriculum learning strategy provides clear benefits. Moreover, adopting the strategy of reconstructing the middle frame from five input frames enhances consistency across reconstructed sequences during inference. Finally, Table~\ref{tab2} reports the detailed performance of CRUNet-MR-Univ at each medical center for both training stages (S1 and S2).

% Requires: \usepackage{graphicx}
\begin{table}[tb]
    \centering
    \caption{Comparison of CRUNet-MR-Univ (proposed) with baseline models on the center-cropped validation set. S1 refers to the first stage, while S2 represents the second stage. Best results are shown in bold.}
    \label{tab1}
    \begin{tabular}{l|ccc}
        \hline
        Methods & PSNR $\uparrow$ & SSIM $\uparrow$ & NMSE $\downarrow$ \\
        \hline
        ZF & 21.765 & 0.584 & 0.113 \\
        SENSE & 23.648 & 0.587 & 0.132 \\
        GRAPPA & 24.433 & 0.639 & 0.084 \\
        CRNN-MRI & 25.900 & 0.730 & 0.076 \\
        UNet-MR & 25.229 & 0.705 & 0.087 \\
        \hline
        CRUNet-MR-Univ w/o CFA \& Prompts (S1) & 26.209 & 0.749 & 0.064 \\
        CRUNet-MR-Univ w/o Prompts (S1) & 26.440 & 0.752 & 0.061 \\
        CRUNet-MR-Univ (S1) & 26.484 & 0.755 & 0.063 \\
        \hline
        CRUNet-MR-Univ (S2) & \textbf{28.232} & \textbf{0.809} & \textbf{0.04} \\
        \hline 
    \end{tabular}
\end{table}

\begin{table}[th]
\centering
\caption{Quantitative multi-center performance evaluation of CRUNet-MR-Univ across two training stages (S1 and S2). Best results are highlighted in bold.}
\label{tab2}
\begin{tabular}{l c c c c c}
\toprule
 & & \multicolumn{2}{c}{CRUNet-MR-Univ (S1)} & \multicolumn{2}{c}{CRUNet-MR-Univ  (S2)} \\
\cmidrule(lr){3-4} \cmidrule(lr){5-6}
Center & Vendor & SSIM$\uparrow$ & PSNR$\uparrow$ & SSIM$\uparrow$ & PSNR$\uparrow$ \\
\hline
C001 & UIH-3.0T-umr780      & 0.737 & 25.80 & \textbf{0.801} & \textbf{27.86} \\
C002 & Siemens-3.0T-CIMA.X  & 0.668 & 23.90 & \textbf{0.727} & \textbf{25.20} \\
C002 & UIH-3.0T-umr880      & 0.727 & 24.86 & \textbf{0.785} & \textbf{27.07} \\
C003 & UIH-3.0T-umr880      & 0.783 & 27.50 & \textbf{0.823} & \textbf{29.32} \\
C004 & Siemens-1.5T-Aera  & 0.710 & 25.24 & \textbf{0.769} & \textbf{27.01} \\
C005 & GE-1.5T-voyager       & 0.808 & 28.62 & \textbf{0.855} & \textbf{30.21} \\
C005 & Siemens-3.0T-Vida  & 0.758 & 26.42 & \textbf{0.820} & \textbf{28.23} \\
C006 & Siemens-3.0T-Prisma  & 0.759 & 27.02 & \textbf{0.816} & \textbf{28.55} \\
C007 & UIH-3.0T-umr790      & 0.814 & 27.84 & \textbf{0.861} & \textbf{29.71} \\
C008 & GE-1.5T-voyager       & 0.788 & 27.63 & \textbf{0.837} & \textbf{29.14} \\
\hline

\multicolumn{2}{c}{Overall Mean} & 0.755 & 26.48 & \textbf{0.809} & \textbf{28.23} \\
\bottomrule
\end{tabular}
\end{table}

\begin{figure}[tb]
\centering
\includegraphics[width=\textwidth]{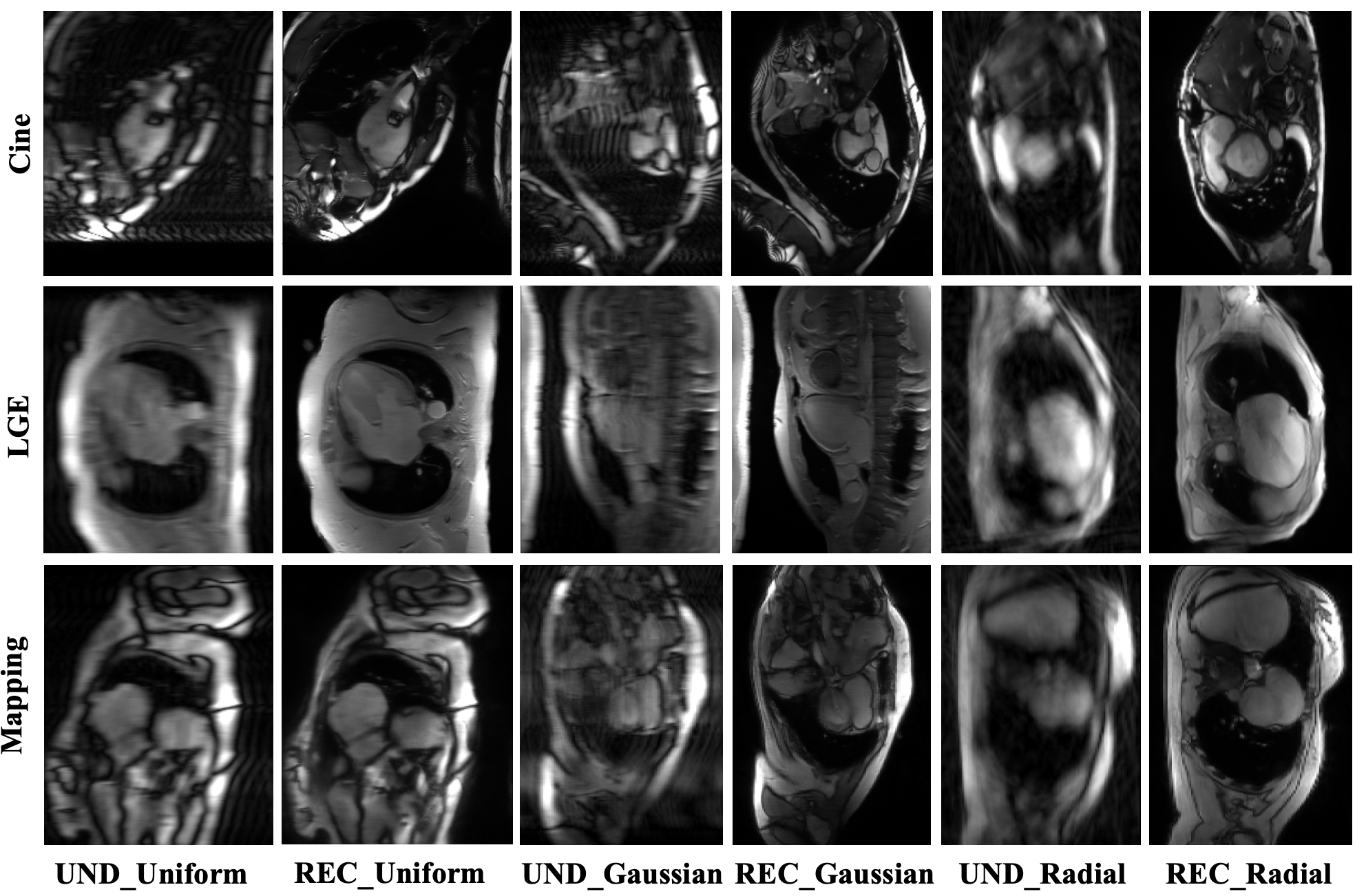}
\caption{Visualizations of CRUNet-MR-Univ (proposed, S2) reconstruction results for three contrasts under three k-space trajectories at an acceleration factor of 24. Here, these cases are from validation set and lack ground truth references. 'REC' indicates the reconstructed images, and 'UND' is the original undersampled inputs.} \label{fig2}
\end{figure}

Figure~\ref{fig2} presents qualitative results of CRUNet-MR-Univ (S2) on randomly selected validation cases, covering multiple contrasts at an acceleration factor of 24 across three sampling trajectories. Overall, the reconstructed results demonstrate effective suppression of aliasing artifacts and significant reduction of blurriness. However, for some certain cases, fine cardiac structures remain partially blurred or insufficiently detailed, indicating that the current model still has room for improvement at high acceleration factor.

\section{Discussion and Conclusion}
As shown in Table~\ref{tab1}, ~\ref{tab2} and Figure~\ref{fig2}, CRUNet-MR-Univ outperforms baseline methods under identical conditions, effectively removing artifacts and recovering fine details across diverse validation scenarios. Introduced components and training strategies have also been shown to positively impact overall reconstruction performance. However, the current performance of CRUNet-MR-Univ on cropped cardiac regions still lags behind the top-ranked methods on the leaderboard. 

Although CRUNet-MR-Univ is trained with additional cascade blocks and more epochs in the second stage, potential limitations in the training process may still affect its performance. Our initial thought is to train the model with more epochs and dynamic learning rate, achieved by setting a relatively small number of training samples per epoch. However, as noted in previous studies~\cite{5,23,24}, extensive epochs are not strictly necessary; rather, ensuring a sufficient number of training samples in each epoch is more important, which aligns with the training principles of foundation models. Another limitation might be the current loss function, which ignores k-space frequency information and relies solely on global image magnitude. Moreover, different imaging contrasts exhibit distinct characteristics, which may also require tailored combinations of loss terms. Designing an improved loss function could allow the model to account for a broader range of reconstruction characteristics. Furthermore, we fix the channel number of each CRUNet block at 64 to reduce GPU memory usage, which may limit the representation of spatial features. In contrast, PromptMR models increase the channel number with depth, enabling richer feature representations. Then, although we leverage an effective combination of convolutional recurrent operations and the U-Net structure to better exploit strong spatio-temporal correlations for reconstruction, the model may still be limited at high acceleration factors due to the restricted receptive field inherent in convolutional operations. In contrast, operations such as applying channel attention along the temporal-channel dimension may offer greater benefits at high acceleration factors by extracting spatio-temporal features within a global receptive field. Therefore, these aspects can be explored in future studies to assess their impact on the overall performance of CRUNet-MR-Univ.

In this work, we propose CRUNet-MR-Univ, a foundation model for CMR reconstruction across diverse conditions. By integrating CRUNet modules, CFA blocks, prompt-based priors, and further refining training process, CRUNet-MR-Univ achieves strong performance and generalization, including on data from unseen medical centers. While some limitations still remain in the current training approach, the model offers strong potential for further improvement.
\\
\\
\textbf{Acknowledgement.} This work is part of the project ROBUST: Trustworthy AI-based Systems for Sustainable Growth with
project number KICH3.LTP.20.006, which is (partly)
financed by the Dutch Research Council (NWO), Philips Research, and the Dutch Ministry of Economic Affairs
and Climate Policy (EZK) under the program LTP KIC
2020-2023.

\bibliographystyle{splncs04}
\bibliography{new}

\newpage
\appendix
\section{Updated Experiment}
Based on discussions about potential future improvements to the current CRUNet-MR-Univ model, we conducted an experiment to further fine-tune the model using a larger number of training samples over fewer epochs, rather than using fewer samples over many epochs. Specifically, the model was trained for 90000 iterations per epoch for a total of four epochs. The evaluation results, presented in Table~\ref{tab3} as Stage 3 (S3), demonstrate the advantage of incorporating more update iterations within each epoch. This observation is also consistent with the training strategies commonly adopted in many large foundation models.

\begin{table}[th]
\centering
\caption{Quantitative multi-center performance evaluation of CRUNet-MR-Univ across three training stages (S1, S2, S3). Best results are highlighted in bold.}
\label{tab3}
\resizebox{1\textwidth}{!}{
\begin{tabular}{l c c c c c c c}
\toprule
 & & \multicolumn{2}{c}{CRUNet-MR-Univ (S1)} & \multicolumn{2}{c}{CRUNet-MR-Univ (S2)} & \multicolumn{2}{c}{CRUNet-MR-Univ (S3)}\\
\cmidrule(lr){3-4} \cmidrule(lr){5-6} \cmidrule(lr){7-8}
Center & Vendor & SSIM$\uparrow$ & PSNR$\uparrow$ & SSIM$\uparrow$ & PSNR$\uparrow$ & SSIM$\uparrow$ & PSNR$\uparrow$ \\
\hline
C001 & UIH-3.0T-umr780      & 0.737 & 25.80 & 0.801 & 27.86 & \textbf{0.809} & \textbf{28.08}\\
C002 & Siemens-3.0T-CIMA.X  & 0.668 & 23.90 & 0.727 & 25.20 & \textbf{0.752} & \textbf{25.77} \\
 C002    & UIH-3.0T-umr880      & 0.727 & 24.86 & 0.785 & 27.07 & \textbf{0.796} & \textbf{27.46} \\
C003 & UIH-3.0T-umr880      & 0.783 & 27.50 & 0.823 & \textbf{29.32} & \textbf{0.830} & 29.15 \\
C004 & Siemens-1.5T-Aera  & 0.710 & 25.24 & 0.769 & 27.01 & \textbf{0.782} & \textbf{27.43} \\
C005 & GE-1.5T-voyager       & 0.808 & 28.62 & 0.855 & 30.21 & \textbf{0.859} & \textbf{30.36} \\
 C005    & Siemens-3.0T-Vida  & 0.758 & 26.42 & 0.820 & 28.23 & \textbf{0.836} & \textbf{28.56} \\
C006 & Siemens-3.0T-Prisma  & 0.759 & 27.02 & 0.816 & 28.55 & \textbf{0.830} & \textbf{29.05} \\
 C006    & UIH-3.0T-umr790      & 0.814 & 27.84 & 0.861 & 29.71 & \textbf{0.867} & \textbf{29.98} \\
C008 & GE-1.5T-voyager       & 0.788 & 27.63 & 0.837 & 29.14 & \textbf{0.842} & \textbf{29.48} \\
\hline

\multicolumn{2}{c}{Overall Mean} & 0.755 & 26.48 & 0.809 & 28.23 & \textbf{0.820} & \textbf{28.530} \\
\bottomrule
\end{tabular}
}
\end{table}

\end{document}